\documentclass[letterpaper, 10 pt, conference]{ieeeconf}  

\IEEEoverridecommandlockouts                              

\newtheorem{theorem}{Theorem}[section]

\newtheorem{assumption}[theorem]{Assumption}
\newtheorem{problem}{Problem}
\newtheorem{rem}[theorem]{Remark}

\usepackage{graphicx} 
\usepackage{amsmath}
\usepackage{amssymb}
\usepackage{epstopdf}
\usepackage{cite}
\usepackage[noend,ruled,linesnumbered]{algorithm2e}
\SetKwComment{Comment}{$\triangleright$\ } {}
\usepackage{multirow}
\usepackage{rotating}
\usepackage{subfigure} 
\usepackage{xcolor}
\usepackage{mysymbol}
\usepackage[dvipsnames]{xcolor}
\usepackage[hyphens]{url}
\usepackage{romannum}
\usepackage{makecell}
\usepackage[breaklinks=true, colorlinks, bookmarks=true, citecolor=Black, urlcolor=Violet,linkcolor=Black]{hyperref}

\usepackage{comment}
\usepackage[colorinlistoftodos,prependcaption,textwidth=1.5cm,textsize=tiny]{todonotes}
\begin{document}

\title{Navigate or Relocate? \\
Planning Among Movable Obstacles in Unknown Environments}

\author{Yuqing Zhang, Haoyu Zhu, Yiannis Kantaros
\thanks{$^{1}$Authors are with the Department of Electrical and Systems Engineering, Washington University in St. Louis, St. Louis, MO, USA. {\tt\small {zyuqing, z.haoyu, ioannisk@wustl.edu}}}}

\maketitle
\begin{abstract}
Conventional robot planning methods seek collision-free paths to a goal but fail when all paths are blocked. In these cases, the robot must determine which objects to relocate, in what order, and where to place them to clear a path---a problem known as Navigation Among Movable Obstacles (NAMO). Most NAMO planners assume a known environment, while existing approaches for unknown environments typically reason locally about relocations and cannot plan interdependent relocation sequences. We consider NAMO in unknown environments revealed through onboard sensing, where the robot must decide whether a blocked route requires relocation or a feasible path may exist through unexplored space. We propose an online framework that addresses this ambiguity by selecting between navigation and relocation using shortest paths that treat discovered movable objects as obstacles or as removable. Navigation relies on existing motion planners, while relocation uses a sampling-based approach that, unlike existing approaches for unknown environments, searches over \textit{interdependent} relocation sequences and uses an LLM to bias sampling. Numerical experiments demonstrate scalability to cluttered environments requiring interdependent relocations and improved plan quality over existing baselines.
\end{abstract}

\maketitle

\section{Introduction}\label{sec:intro}
\vspace{-0.1cm}

Robot navigation seeks to compute collision-free paths from an initial state to a goal region \cite{karaman2011sampling}. 
When obstacles block all such paths, a robot may instead relocate movable objects, requiring it to determine which objects to move, in what order, and where to place them. This gives rise to the Navigation Among Movable Obstacles (NAMO) problem \cite{wilfong1988motion}, which has been studied using heuristic, search-based, and sampling-based approaches 
\cite{stilman2005navigation,stilman2008planning,nieuwenhuisen2008effective,van2010path,moghaddam2016planning,bayraktar2023solving,kalluraya2026making,11204512}. These works assume the environment is known a priori, allowing offline planning.

\begin{figure}[t]
\centering
    \includegraphics[width=\linewidth]{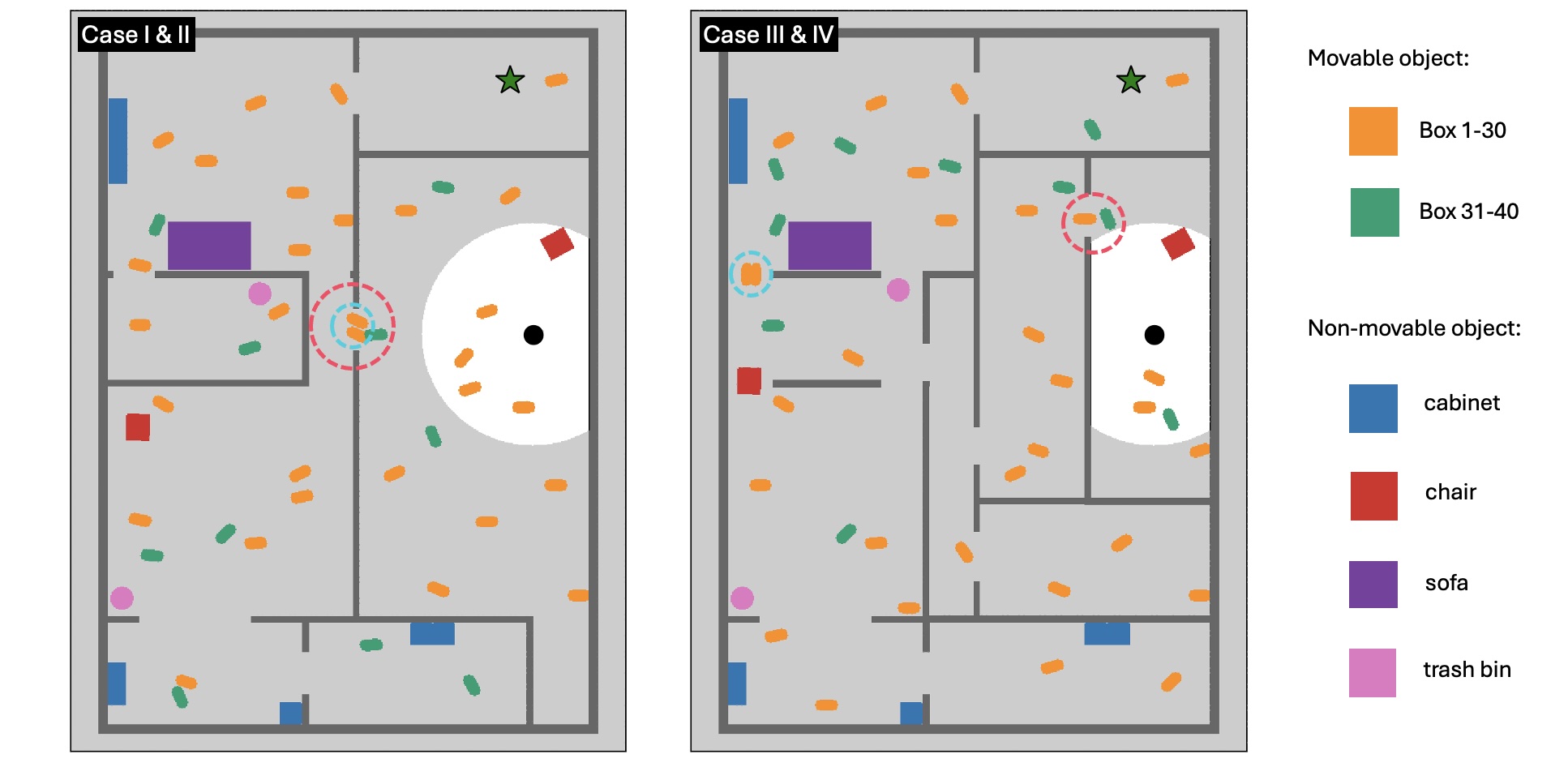}
    \vspace{-0.75cm}
    \caption{Case-study environments of Section~\ref{sec:sims}: Cases I–II (left) and III–IV (right). The robot (black disk) must reach the goal (green star) in an  unknown environment with non-movable furniture and movable boxes; green boxes are added in Cases II and IV. Only the sensed region (white) is known; the rest (gray) is unobserved. Cyan and red circles mark passages requiring joint relocations and resolving access dependencies, respectively.}
    \label{fig:intro}
    \vspace{-0.1cm}
\end{figure}

In this work, we consider NAMO in an unknown environment, where the robot discovers the environment structure (e.g., walls) and the locations and semantic labels of objects through onboard perception. The goal may be unreachable in known free space because discovered movable
objects require relocation or a collision-free path may exist through unobserved regions; see Fig.~\ref{fig:intro}.
The robot must therefore decide whether to navigate to the goal or relocate discovered objects. We address this problem using an online planning framework that interleaves mapping, decision-making, and execution. 
Given a map constructed using existing methods
\cite{hughes2022hydra}, our framework compares two shortest paths on optimistic versions of the map: one treats discovered movable objects as obstacles and the other as removable.
Relocation is selected only if the latter reaches the goal within known space or enters unknown space through a frontier inaccessible to the former; otherwise, the robot continues navigating.
Navigation relies on existing motion planning methods, while our $\text{NAMO-LLM}_{\text{u}}$ extends the sampling-based NAMO-LLM planner \cite{11204512} to partially known environments.
It builds trees over hypothetical relocation sequences of discovered objects, using a pre-trained LLM to bias sampling toward promising directions. Compared to NAMO-LLM, our method introduces a termination criterion that accounts for access to unexplored space, a tree representation that avoids explicitly computing free-space components (which is
computationally expensive), and a sampling strategy and LLM prompt tailored to partial environment knowledge. Finally, following \cite{van2010path,11204512}, we establish its probabilistic completeness.

\textbf{Related Works:} \textit{(i) NAMO in Unknown Environments.}
The works most closely related to ours address NAMO in unknown environments
\cite{wu2010navigation,levihn2014locally,ellis2022object,armleder2024tactile,he2024interactive,kakiuchi2010working,stan2025adaptive}.
The method in \cite{wu2010navigation} builds a grid map online, treats unexplored space as free, and, when the current path is blocked, chooses based on estimated costs between navigating around obstacles and pushing a single object. 
Related works improve efficiency \cite{levihn2014locally}, demonstrate this approach on real robots \cite{ellis2022object}, or infer movability through contact \cite{armleder2024tactile}.
The method in \cite{he2024interactive} encodes per-object pushing costs in a sparse visibility graph built online and likewise trades off pushing against detouring. Other approaches consider the same setting without explicit relocation planning \cite{kakiuchi2010working,stan2025adaptive}; for example, \cite{stan2025adaptive} lets the robot drive into a priori unknown obstacles and reroutes when they do not give way. These approaches reason locally about individual object relocations and do not plan interdependent relocation sequences, e.g., when one object must first be moved to access another or create space for its relocation. Our framework plans such sequences and decides when to relocate objects based on whether relocation opens a route inaccessible through navigation alone rather than by comparing navigation and manipulation costs with manually set relative weights \cite{wu2010navigation,levihn2014locally,ellis2022object,armleder2024tactile,he2024interactive}.
The visibility-aware algorithm in \cite{muguira2023visibility} likewise
assumes independent object relocations.
Our method instead searches over interdependent relocation sequences where, e.g., relocating one object is required to make another relocation feasible.

The method in \cite{ellis2023navigation} extends the above works to dependent relocations. It selects a relocation only if it opens a path to the goal or exposes a previously unreachable manipulation point of another blocking object. This handles access dependencies, in which one object blocks access to the manipulation point of another, but neither space dependencies, in which one object occupies the space another must be moved into, nor joint relocations, where several reachable objects must be moved before any path opens; see Fig. \ref{fig:intro}. Our method handles these cases by searching over complete relocation sequences. Since several assumptions in \cite{ellis2023navigation} do not hold in our setting, we compare against a baseline that builds upon this work; see Section~\ref{sec:sims}.
Another approach that can handle certain relocation dependencies is the reactive architecture of \cite{vasilopoulos2021reactive}. 
Its setting imposes additional structural assumptions, including geometric familiarity or convexity of fixed obstacles and obstacle-separation conditions.
It sequentially moves objects that block the manipulation of other objects, but, unlike our work, uses heuristic placements rather than explicitly planning relocation sequences and configurations in which the placement of one object must enable the subsequent relocation of another. 

Learning-based approaches have also been proposed \cite{zeng2021pushing,wang2023curriculum,yao2024local,zhou2025adaptive}. These methods learn policies for selecting navigation and object-interaction actions, but do not verify the long-horizon feasibility of the resulting interactions. For example, \cite{yao2024local} reports failures in which an object is pushed to a configuration that irreversibly blocks the path to the goal. In contrast, our planner explicitly searches over complete relocation sequences and configurations and verifies their feasibility, ensuring the soundness of returned plans. These learned policies are complementary to our approach and could, for example, replace the LLM in biasing the sampling of candidate relocations.
 
\textit{(ii) Rearrangement Planning.} Our work is related to rearrangement planning (RP), where objects are moved to prescribed arrangements \cite{labbe2020monte,krontiris2015dealing,11128108,ren2022rearrangement,wang2022efficient}. 
However, RP differs from NAMO as it considers predefined goal positions for movable objects and typically does not specify a final robot goal.

\textbf{Summary of Contributions:} First, we propose an online NAMO planner for unknown environments that plans complete relocation sequences, including interdependent and joint relocations. Second, we develop $\text{NAMO-LLM}_\text{u}$, a probabilistically complete sampling-based planner that extends NAMO-LLM from known to partially known environments. 
Third, experiments demonstrate scalability to cluttered environments requiring interdependent and joint relocations, as well as improved plan quality over baseline methods. 


\vspace{-0.15cm}
\section{Problem Formulation}\label{sec:pf}
\vspace{-0.1cm}

\textbf{Robot Modeling:} We consider a robot with dynamics $\bbp(t+1)=\bbf(\bbp(t),\bbu(t))$, where $\bbp(t)\in\ccalP$ and $\bbu(t)\in\ccalU$ denote its configuration and control input. We assume that $\bbp(t)$ is known and that the robot is holonomic and can track any collision-free path in $\ccalP$. The robot has a gripper and can manipulate one movable object at a time, subject to the following assumption.

\begin{assumption}[Carried-object footprint]\label{as:carry}While carrying an object, the robot--object system has the same footprint, and thus collision constraints, as the robot alone.
\end{assumption}

\textbf{Environment Modeling:} The robot operates in a bounded workspace $\ccalW\subset\mathbb{R}^2$ that is unknown except for its outer boundary. The workspace contains rigid obstacles $\ccalO=\{o_1,\dots,o_N\}$, whose number, shapes, locations, and semantic classes are initially unknown. Each obstacle belongs to a semantic class $\ell_i\in\ccalL$ and is \textit{movable} iff $\ell_i\in\ccalL^{\mathrm{mov}}\subseteq\ccalL$, where $\ccalL^{\mathrm{mov}}$ is known. We denote the set of movable obstacles by $\ccalO^{\mathrm{mov}}$ and the configuration of each movable object $o_i$ by $\bbc_i(t)\in\ccalC_i$, where $\ccalC_i$ is its configuration space.

\textbf{Sensing and Mapping:} The robot uses an onboard, range-limited RGB-D sensor to construct a metric-semantic map online. Its occupancy layer discretizes $\ccalW$ into a grid graph $\ccalG=(\ccalV,\ccalE)$, known a priori from the workspace boundary and fixed grid resolution. At time $t$, the cells are partitioned into known-free, known-occupied, and unknown sets, $\ccalV_t^f$, $\ccalV_t^o$, and $\ccalV_t^u$, respectively. An object layer maintains the discovered movable objects $\ccalO^{\mathrm{mov}}(t)\subseteq\ccalO^{\mathrm{mov}}$, their configurations $\bbc_i(t)$, and footprints $\ccalB_i(\bbc_i(t))\subseteq\ccalV$. Both layers are updated online from sensor observations and executed relocations. This representation can be obtained from existing metric-semantic mapping systems such as Hydra \cite{hughes2022hydra}; Section~\ref{sec:sims} describes our implementation.
We model the robot as a point and, with a slight abuse of notation, write $\bbp(t)\in\ccalV$ for its occupied cell.\footnote{In practice, obstacles are inflated by the robot radius.} We also denote by $\ccalN_t\subseteq\ccalV_t^f$ the cells reachable from $\bbp(t)$ through known-free cells. 
As in prior related work, we make the following assumption about the constructed map \cite{wu2010navigation,levihn2014locally,ellis2023navigation,muguira2023visibility,he2024interactive,zhou2025adaptive}.

\begin{assumption}[Map correctness]\label{as:perception}
The map information available to the robot is assumed to be correct for all $t$. 
\end{assumption}

\textbf{Robot Task:} The robot must reach a goal region, given a priori as a set of cells $\ccalV_g\subseteq \ccalV$, from an initial position $\bbp(0)\in\ccalV$ while avoiding obstacles.  We consider cases where $\ccalV_g\cap\ccalN_0=\emptyset$, either because movable objects block all known paths to the goal or because a feasible path traverses unexplored space. The robot must therefore decide online whether to continue toward the goal, passively revealing the map, or relocate discovered objects and, if so, which objects to relocate, in what order, and where.

\begin{figure}[t]
\centering
\includegraphics[width=0.7\linewidth]{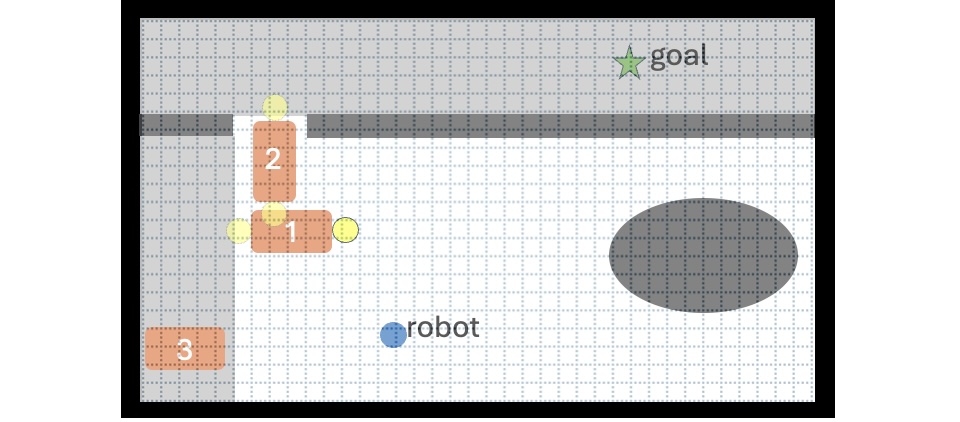}
\vspace{-0.4cm}
\caption{Illustration of object categories and task setting. The workspace is discretized into a grid graph $\ccalG$ (dotted lines); white and light gray denote known-free and unexplored space, respectively. Boxes 1--3 are movable: box 1 is manipulable because one manipulation point (yellow dot) is reachable; box 2 is discovered but non-manipulable since its manipulation points are blocked or lie in the unexplored space; and box 3 is undiscovered. Reaching the goal requires relocating box 1 before box 2.
} 
\label{fig:objcat}
\vspace{-0.1cm}
\end{figure}

\textbf{Manipulable Objects:} Each movable object $o_i$ has $J_i\in\mathbb{N}_{>0}$ known manipulation locations fixed in its local frame. Let $m^j(\bbc_i)\in\ccalV$ denote the cell the robot must occupy to grasp or release $o_i$ through its $j$-th manipulation location at configuration $\bbc_i\in\ccalC_i$, and let $\ccalM(\bbc_i):=\{m^j(\bbc_i)\}_{j=1}^{J_i}$ be the set collecting the manipulation cells of $o_i$ at $\bbc_i$. The set of \textit{manipulable} objects at time $t$ is
\begin{equation}\label{eq:manipulable}
    \ccalO^{M}(t):=
    \left\{
        o_i\in \ccalO^{\mathrm{mov}}(t)
        \,\middle|\,
        \ccalM(\bbc_i(t))\cap\ccalN_t\neq\emptyset
    \right\},
\end{equation}
i.e., the discovered movable objects with at least one reachable manipulation cell; see Fig.~\ref{fig:objcat}.

\textbf{Robot Primitives:} The robot executes \emph{waypoint transitions} and \emph{carry actions} using existing low-level controllers. A waypoint transition moves the robot, while not carrying an object, to $w\in\ccalV$ through known-free space and is feasible if $w\in\ccalN_t$. Newly sensed cells are passively revealed during execution.
A carry action $a^{j}_{\mathrm{car}}(\bbc_i(t),\bbc_i')$ grasps $o_i$ at $m^j(\bbc_i(t))$, carries it, and releases it at $m^j(\bbc_i')$. By Assumption~\ref{as:carry}, carrying an object does not alter the collision constraints imposed on the robot. A carry action is feasible at time $t$ if (i) $\bbp(t)=m^j(\bbc_i(t))$; (ii)
$\ccalB_i(\bbc_i')\subseteq\ccalV_t^f\cup\ccalB_i(\bbc_i(t))$; and (iii) the robot can move from $m^j(\bbc_i(t))$ to $m^j(\bbc_i')$ through cells in $\ccalN_t\cup\ccalB_i(\bbc_i(t))$. Thus, approaching an object is done by waypoint transitions, whereas grasping, transporting, and releasing it are captured by a single
carry action.

\textbf{Robot Plan:} A robot plan $\tau=(a_1,\dots,a_L)$ is a finite sequence of waypoint transitions and carry actions and is feasible if every action is feasible when executed.

\begin{problem}\label{pr1}
Given an initial robot position $\bbp(0)$ and a goal region $\ccalV_g\subseteq\ccalV$ such that $\ccalV_g\cap\ccalN_0=\emptyset$, compute online a feasible robot plan $\tau$ that drives the robot to $\ccalV_g$.
\end{problem} 

\vspace{-0.1cm}
\section{Proposed Planning Algorithm}\label{sec:method}
\vspace{-0.1cm}

In this section, we present the proposed framework for Problem~\ref{pr1}; see Alg.~\ref{alg:framework}. Since the environment is revealed online, the framework constructs the plan $\tau$ incrementally through a receding-horizon loop that interleaves sensing, decision-making, and execution. 
Section~\ref{sec:decision} presents the mechanism that arbitrates, at each decision instant, between navigation and object relocation, and describes the execution of the selected mode. The navigation mode relies on existing planning algorithms (e.g., $A^\star$), while the relocation mode uses a new sampling-based NAMO planner, called $\text{NAMO-LLM}_{\text{u}}$, which builds upon \cite{11204512}. Section~\ref{sec:tree} presents the internal design of $\text{NAMO-LLM}_{\text{u}}$. 

\begin{algorithm}[t]
\footnotesize
\caption{Proposed Planning Framework}
\LinesNumbered
\label{alg:framework}
\KwIn{Initial robot configuration $\bbp(0)$; Goal region $\ccalV_g$}
\KwOut{plan $\tau$ with $\bbp(T)\in\ccalV_g$, or \textsc{Fail}}
$k\gets 0$;~ $t\gets t_0\gets 0$;~ $\tau\gets\emptyset$\; \label{alg1:init}
\While{$\bbp(t)\notin\ccalV_g$}{ \label{alg1:while}
Update the map $\ccalG$ and the stacked configurations $\bbc(t)$\; \label{alg1:map}
Compute $\widehat{\ccalO}^{M}(t)$ and construct $\ccalG_t^{\text{nav}}$, $\ccalG_t^{\text{man}}$\; \label{alg1:closure}
$[\pi_t^{\text{nav}},J_t^{\text{nav}}]\gets A^\star\big(\bbp(t),\ccalV_g;~\ccalG_t^{\text{nav}}\big)$\; \label{alg1:nav}
$[\pi_t^{\text{man}},J_t^{\text{man}},n_t]\gets A^\star\big(\bbp(t),\ccalV_g;~\ccalG_t^{\text{man}}\big)$\; \label{alg1:man}
\If{$J_t^{\text{man}}=\infty$}{\Return \textsc{Fail}\; \label{alg1:fail}}
Compute $\mathrm{match}_t$\;\label{alg1:match}
\eIf{$\neg\,\mathrm{match}_t$}{ \label{alg1:branch}
$(a_1,\dots,a_h)\gets$ Alg.~\ref{alg:planner}$\big(\bbp(t),\bbc(t),\ccalG,\pi_t^{\text{nav}},\pi_t^{\text{man}},\ccalV_g\big)$\; \label{alg1:invoke}
Navigate to the grasp cell of $a_1$, then execute $a_1$;~ append the traversed segment and $a_1$ to $\tau$\; \label{alg1:execman}
}{
Follow $\pi_t^{\text{nav}}$ for up to $\delta$ steps; append the traversed prefix to $\tau$\;  \label{alg1:explore}
}
$k\gets k+1$;~ $t\gets t_k$, the end time of the executed segment\; \label{alg1:tick}
}
\Return $\tau$\; \label{alg1:return}
\end{algorithm}

\vspace{-0.2cm}
\subsection{Deciding Between Navigation and Manipulation}\label{sec:decision}
\vspace{-0.1cm}

Each iteration of the proposed method starts at a \emph{decision instant} $t$, which need not occur at consecutive time steps.
The robot first updates its map $\ccalG$ and the configurations of discovered objects [line~\ref{alg1:map}, Alg.~\ref{alg:framework}]. It then selects between navigation and relocation using two shortest-path queries that differ in whether discovered movable objects are treated as obstacles or as if they were absent.

\textbf{Graph Structures:} To define these queries, we introduce two graphs constructed from $\ccalG$. The \textit{detour graph} $\ccalG_t^{\text{nav}}$ optimistically treats all unknown cells as free; thus, its node set is $\ccalV_t^f\cup\ccalV_t^u$. Two nodes are connected if their cells share at least one point.
To define the second graph, observe that relocations can cascade: relocating a currently manipulable object may make the manipulation cells of other movable objects reachable. For example, in Fig.~\ref{fig:objcat}, box~2 becomes manipulable only once box~1 is relocated. We therefore define the \emph{recursively manipulable} set $\widehat{\ccalO}^{M}(t)$ as the fixed point of the following monotone recursion: initialize $\ccalD^0=\ccalO^{M}(t)$ and, given $\ccalD^k$, let $\ccalD^{k+1}$ collect every $o_i\in\ccalO^{\mathrm{mov}}(t)$ with at least one manipulation cell in $\ccalM(\bbc_i(t))$ that is reachable from $\bbp(t)$ through cells in $\ccalV_t^f\cup\bigcup_{o_l\in\ccalD^k}\ccalB_l(\bbc_l(t))$, i.e., through known-free cells and the footprints of the objects in $\ccalD^k$, treated as vacated. In other words, $\ccalD^{k+1}$ evaluates \eqref{eq:manipulable} with $\ccalN_t$ replaced by the set of cells reachable from $\bbp(t)$ once the objects in $\ccalD^k$ have been removed. Since removing objects can only enlarge this set, $\ccalD^k\subseteq\ccalD^{k+1}\subseteq\ccalO^{\mathrm{mov}}(t)$, and the fixed point is reached in at most $|\ccalO^{\mathrm{mov}}(t)|$ iterations [line~\ref{alg1:closure}, Alg.~\ref{alg:framework}]. The recursion runs on known space: reachability is evaluated over known-free cells and freed footprints, never through unknown cells.
Given $\widehat{\ccalO}^{M}(t)$, we define the \textit{clearing graph} $\ccalG_t^{\text{man}}$ by treating all unknown cells as free and objects in $\widehat{\ccalO}^{M}(t)$ as absent. Its node set is $\ccalV_t^f\cup\ccalV_t^u\cup\bigcup_{o_i\in\widehat{\ccalO}^{M}(t)}\ccalB_i(\bbc_i(t))$, with edges defined as in $\ccalG_t^{\text{nav}}$.

\begin{figure}[t]
\centering
\includegraphics[width=\linewidth]{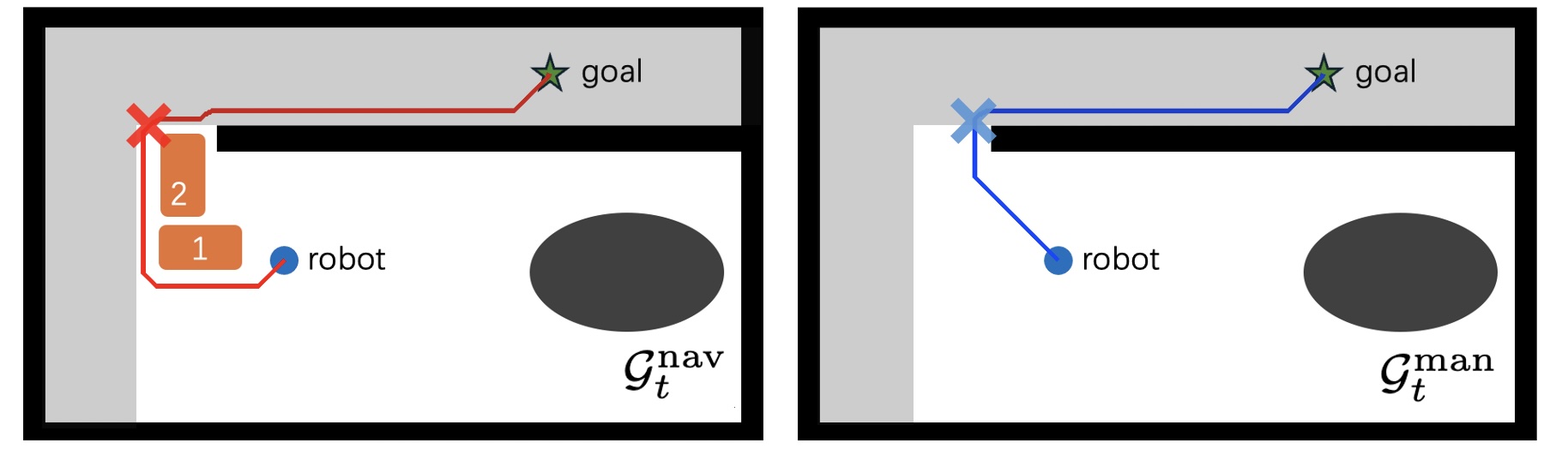}
\vspace{-0.7cm}
\caption{Mode selection in \eqref{eq:decision}, illustrated in the environment of Fig.~\ref{fig:objcat}. Unknown space (light gray) contains the goal (green star). The detour path $\pi_t^{\text{nav}}$ (red) and clearing path $\pi_t^{\text{man}}$ (black) are computed on the optimistic graphs $\ccalG_t^{\text{nav}}$ and $\ccalG_t^{\text{man}}$, which treat unknown
cells as traversable. Boxes 1 and 2 obstruct the passage in
$\ccalG_t^{\text{nav}}$, so the paths enter unknown space at different frontier cells (crosses); thus, $\phi(\pi_t^{\text{nav}})\neq\phi(\pi_t^{\text{man}})$ and
\textsc{Relocate} is selected.
}
\label{fig:modes}
\vspace{-0.2cm}
\end{figure}

\textbf{Shortest-path Queries:} The first query computes the \emph{detour path} $\pi_t^{\text{nav}}$ by running $A^\star$ (or any shortest-path algorithm) over $\ccalG_t^{\text{nav}}$; it is the best route to the goal that relocates no discovered object. The second computes the \emph{clearing path} $\pi_t^{\text{man}}$ over $\ccalG_t^{\text{man}}$, which may traverse the footprints of recursively manipulable objects as if they were absent; let $n_t$ denote the number of distinct objects in $\widehat{\ccalO}^{M}(t)$ whose footprints intersect $\pi_t^{\text{man}}$. The path lengths are denoted by $J_t^{\text{nav}}$ and $J_t^{\text{man}}$, each set to $\infty$ if the corresponding path does not exist [lines~\ref{alg1:nav}--\ref{alg1:man}, Alg.~\ref{alg:framework}]. Since $\ccalG_t^{\text{nav}}$ is a subgraph of $\ccalG_t^{\text{man}}$, $J_t^{\text{man}}\leq J_t^{\text{nav}}$; thus, the clearing path exists whenever the detour path does. Moreover, if $n_t=0$, then $J_t^{\text{man}}=J_t^{\text{nav}}$. These paths are visualized in Fig. \ref{fig:modes}.

\textbf{Frontier matching:} For a path $\pi$, we define its \emph{frontier} $\phi(\pi)$ as the first cell of $\pi$ that lies in $\ccalV_t^u$, i.e., the first cell through which $\pi$ leaves known space, and set $\phi(\pi)=\bot$ if $\pi$ never enters unknown space. Up to its frontier, a path (detour or clearing) lies in known space and is therefore collision-free given the current map (with the clearing path requiring the relocation of any objects it crosses); beyond its frontier, the path relies on the optimistic assumption that unknown cells are free.
We say that the two paths \emph{match}, and write $\mathrm{match}_t=\text{true}$, if the detour path exists and at least one of the following holds [line~\ref{alg1:match}, Alg.~\ref{alg:framework}]:
\begin{enumerate}
    \item[(i)] $n_t=0$: the clearing path crosses no object and is therefore
    itself a shortest detour path;
    \item[(ii)] $\phi(\pi_t^{\text{nav}})=\bot$: the detour path reaches the goal within known space;
    \item[(iii)] $\phi(\pi_t^{\text{nav}})=\phi(\pi_t^{\text{man}})$: the two paths leave known space through the same cell.
\end{enumerate}
Otherwise, $\mathrm{match}_t=\text{false}$. Note that the paths never match when the detour path does not exist and that $\mathrm{match}_t=\text{true}$ does not imply that the two paths coincide geometrically.

\textbf{The decision:} The mode at a decision instant $t$ is selected by the following cascade [lines~\ref{alg1:fail}--\ref{alg1:explore}, Alg.~\ref{alg:framework}]:
\begin{equation}\label{eq:decision}
    \mathrm{mode}_t=
    \begin{cases}
        \textsc{Fail}, & \text{if } J_t^{\text{man}}=\infty,\\
        \textsc{Relocate}, & \text{else if } \neg\,\mathrm{match}_t,\\
        \textsc{Navigate}, & \text{otherwise}.
    \end{cases}
\end{equation}

\textsc{Fail} is declared when neither path exists ($J_t^{\text{man}}=\infty$, which implies $J_t^{\text{nav}}=\infty$): the goal is disconnected from the robot even when all unknown cells are treated as free and all recursively manipulable objects as removed. Since relocating discovered objects cannot make any object outside $\widehat{\ccalO}^{M}(t)$ manipulable, no relocation sequence can open a route to the goal, and the problem is therefore infeasible on the current map [line~\ref{alg1:fail}, Alg.~\ref{alg:framework}].
\textsc{Relocate} is selected when $\mathrm{match}_t=\text{false}$, i.e., when the clearing path exploits object relocation to reach the goal within known space or enter unknown space through a different frontier cell than the detour path, or when no detour path exists. The framework then invokes Alg.~\ref{alg:planner} to compute a relocation sequence, whose first action is executed [lines~\ref{alg1:invoke}--\ref{alg1:execman}, Alg.~\ref{alg:framework}].
\textsc{Navigate} is selected when $\mathrm{match}_t=\text{true}$. 
The two operating modes are illustrated in Fig.~\ref{fig:modes}.

\textbf{Designing Relocation Sequences:}
In \textsc{Relocate}, the framework invokes the sampling-based planner of Alg.~\ref{alg:planner}. Given the current map $\ccalG$, robot cell $\bbp(t)$, object configurations $\bbc(t)$, goal region $\ccalV_g$, and paths $\pi_t^{\text{nav}}$ and $\pi_t^{\text{man}}$, the planner searches for a feasible sequence $(a_1,\dots,a_h)$ of carry actions such that, after execution, the detour path from the resulting robot cell either reaches the goal within known space or leaves known space through the frontier $\phi(\pi_t^{\text{man}})$ of the clearing path [line~\ref{alg1:invoke}, Alg.~\ref{alg:framework}]. All placements and transport paths in the returned sequence lie in known free space; feasibility conditions are detailed in Section~\ref{sec:tree}.

\textbf{Designing Navigation Paths:} In \textsc{Navigate}, the framework uses the detour path $\pi_t^{\text{nav}}$, computed by $A^\star$ over $\ccalG_t^{\text{nav}}$, as the navigation path.

\textbf{Executing Relocation Sequences:}
The framework executes only the first action $a_1$ of the returned sequence before replanning [line~\ref{alg1:execman}, Alg.~\ref{alg:framework}]. The robot first navigates within $\ccalN_t$ to the grasp cell $m^j(\bbc_i(t))$ of $a_1$, establishing condition (i) of Section~\ref{sec:pf}, and then carries $o_i$ to the release cell $m^j(\bbc_i')$, treating its vacated cells as free. The paths for both stages are computed by Alg.~\ref{alg:expand} when certifying the feasibility of $a_1$ (see Section~\ref{sec:tree}). The map is updated during execution, but these updates cannot invalidate the planned paths: by Assumption~\ref{as:perception}, sensing only relabels unknown cells, whereas $a_1$ uses only known free space. Thus, $a_1$ is completed before the relocation-navigation decision is re-evaluated according to \eqref{eq:decision} [line~\ref{alg1:tick}, Alg.~\ref{alg:framework}].

\begin{rem}[Relocation Sequence]
The unexecuted suffix $(a_2,\dots,a_h)$ can be reused at the next call, provided it remains feasible on the updated map and still resolves the mismatch; our implementation in Section~\ref{sec:sims} does so. This reduces replanning cost without altering the framework.
\end{rem}

\textbf{Executing Navigation Paths:}
In \textsc{Navigate}, the robot executes a prefix of $\pi_t^{\text{nav}}$ for at most $\delta$ time steps, while  updating the map, and appends the traversed prefix to $\tau$ [line~\ref{alg1:explore}, Alg.~\ref{alg:framework}]. 
Execution terminates earlier if newly sensed obstacles block the path. Upon termination, the relocation-navigation decision is re-evaluated as in \eqref{eq:decision} [lines~\ref{alg1:map}--\ref{alg1:branch}, Alg.~\ref{alg:framework}].

\textbf{Construction of Plan $\tau$:} The executed waypoint transitions and carry actions are appended to $\tau$ in order of execution [lines~\ref{alg1:execman} and \ref{alg1:explore}, Alg.~\ref{alg:framework}]. Since every action is feasible when executed and the loop terminates only when $\bbp(t)\in\ccalV_g$ [line~\ref{alg1:while}, Alg.~\ref{alg:framework}], the returned $\tau$ is by construction a feasible plan that solves Problem~\ref{pr1} [line~\ref{alg1:return}, Alg.~\ref{alg:framework}].

\vspace{-0.1cm}
\subsection{Sampling-based Planning of Relocation Sequences}\label{sec:tree} 
\vspace{-0.1cm}

The planner invoked in \textsc{Relocate}, termed $\text{NAMO-LLM}_{\text{u}}$, builds upon NAMO-LLM \cite{11204512}, which computes relocation sequences in known environments by growing a tree of hypothetical relocation sequences. As discussed in Section~\ref{sec:intro}, $\text{NAMO-LLM}_{\text{u}}$ introduces three main modifications: (i) a new termination criterion for partially known environments; (ii) a tree representation that avoids computing free-space components (which is often computationally expensive) by explicitly tracking the robot and object configurations; and (iii) a modified relocation-sampling strategy that restricts hypothetical relocations to known free space and favors nearby object placements to reduce relocation travel.
In what follows, we describe $\text{NAMO-LLM}_{\text{u}}$ in detail.

\textbf{Tree Representation and Initialization:} At each invocation, the current map $\ccalG$ and robot and object configurations $\bbp(t)$ and $\bbc(t)$ are frozen, defining a \emph{snapshot} on which the tree is constructed. The frontier of the clearing path $\pi_t^{\text{man}}$ computed for \eqref{eq:decision} is stored as $z_{\mathrm{ref}}=\phi(\pi_t^{\text{man}})$ for the termination test described below [line~\ref{alg2:freeze}, Alg.~\ref{alg:planner}].
The planner grows a tree $\ccalT=(\ccalV_{\ccalT},\ccalE_{\ccalT})$, where $\ccalV_{\ccalT}$ and $\ccalE_{\ccalT}$ denote its sets of nodes and edges. Each node $\bbq=[\bbp,\bbc]$ represents hypothetical robot and object configurations resulting from the relocations along the corresponding branch, while each edge $(\bbq,\bbq')$ represents a single carry action, recorded by a function $A$, that transitions from $\bbq$ to $\bbq'$. 
Each node $\bbq$ induces a \emph{simulated map}, obtained from the snapshot by placing the objects according to $\bbc$, and an associated detour path $\pi^{\text{nav}}(\bbq)$, computed by $A^\star$ from $\bbp$ on this simulated map. We define $z(\bbq)$ as the frontier of this path, i.e., $z(\bbq)=\phi(\pi^{\text{nav}}(\bbq))$, with $z(\bbq)=\bot$ if the path avoids unknown space and $z(\bbq)=\varnothing$ if no such path exists. Thus, $z(\bbq)\in\ccalV_t^u\cup\{\bot,\varnothing\}$. 
Unlike the fixed $z_{\mathrm{ref}}$, $z(\bbq)$ depends on the node and is
computed when the node is added to the tree for use in the termination test.
%
The tree is initialized as $\ccalV_{\ccalT}=\{\bbq_0\}$ and $\ccalE_{\ccalT}=\emptyset$, where the root $\bbq_0=[\bbp(t),\bbc(t)]$ represents the snapshot [lines~\ref{alg2:root}--\ref{alg2:init}, Alg.~\ref{alg:planner}]. By construction, $z(\bbq_0)=\phi(\pi_t^{\text{nav}})$ if $\pi_t^{\text{nav}}$ exists, and $z(\bbq_0)=\varnothing$ otherwise.

\begin{algorithm}[t]
\footnotesize
\caption{$\text{NAMO-LLM}_{\text{u}}$}
\LinesNumbered
\label{alg:planner}
\KwIn{$\bbp(t)$; $\bbc(t)$, map $\ccalG$; $\pi_t^{\text{nav}},\pi_t^{\text{man}}$; $\ccalV_g$}
\KwOut{relocation sequence $(a_1,\dots,a_h)$}
Freeze the current information;~ $z_{\text{ref}}\gets\phi(\pi_t^{\text{man}})$\; \label{alg2:freeze}
$\textcolor{black}{\bbq_0}\gets\textcolor{black}{[\bbp(t),\bbc(t)]}$\; \label{alg2:root}
$\ccalV_{\ccalT}\gets\{\textcolor{black}{\bbq_0}\}$,~ $\ccalE_{\ccalT}\gets\emptyset$\; \label{alg2:init}
\While{\textnormal{true}}{ \label{alg2:while}
Sample a node $\bbq_{\text{rand}}\in\ccalV_{\ccalT}$ using $f_{\ccalT}$\; \label{alg2:sample}
$[\bbq_{\text{new}},\textcolor{black}{a^{j}_{\mathrm{car}}(\bbc_i,\bbc_i')}]\gets\texttt{TreeExpansion}(\bbq_{\text{rand}})$\; \label{alg2:expand}
\If{$\bbq_{\text{new}}\neq\varnothing$}{ \label{alg2:if}
$\ccalV_{\ccalT}\gets\ccalV_{\ccalT}\cup\{\bbq_{\text{new}}\}$,~ $\ccalE_{\ccalT}\gets\ccalE_{\ccalT}\cup\{(\bbq_{\text{rand}},\bbq_{\text{new}})\}$,~ $A((\bbq_{\text{rand}},\bbq_{\text{new}}))\gets \textcolor{black}{a^{j}_{\mathrm{car}}(\bbc_i,\bbc_i')}$\; \label{alg2:add}
\textcolor{black}{Compute $\pi^{\text{nav}}(\bbq_{\text{new}})$ and $z(\bbq_{\text{new}})$}\; \label{alg2:frontier}
\If{\textcolor{black}{$z(\bbq_{\text{new}})\in\{z_{\text{ref}},\bot\}$}}{ \label{alg2:target}
$\bbq^\star\gets\bbq_{\text{new}}$;~ \Return $(a_1,\dots,a_h)$, extracted by backtracking from $\textcolor{black}{\bbq_0}$ to $\bbq^\star$ via $A$\; \label{alg2:extract}
}
}
}
\end{algorithm}

\textbf{Tree Expansion:} At each iteration, the planner samples a node $\bbq_{\text{rand}}=[\bbp_{\text{rand}},\bbc_{\text{rand}}]\in \ccalV_{\ccalT}$ according to [line~\ref{alg2:sample}, Alg.~\ref{alg:planner}]
\begin{equation}\label{eq:fv}
    f_{\ccalT}
    =p_{\text{rand}}\,\mathrm{Unif}(\ccalV_{\ccalT}^{\max})+
    (1-p_{\text{rand}})\,\mathrm{Unif}(\ccalV_{\ccalT}),
\end{equation}
where $\mathrm{Unif}(\cdot)$ denotes the uniform distribution over a set, $p_{\text{rand}}\in[0,1)$ is user-specified, and $\ccalV_{\ccalT}^{\max}\subseteq\ccalV_{\ccalT}$ contains the nodes identified as most promising for expansion. The construction of $\ccalV_{\ccalT}^{\max}$ is detailed in the Appendix. 
The sampled node is then expanded using Alg.~\ref{alg:expand} [line~\ref{alg2:expand}, Alg.~\ref{alg:planner}]. 

Alg.~\ref{alg:expand} first computes the manipulable set $\ccalO^{M}_{\text{rand}}$ on the simulated map induced by $\bbq_{\text{rand}}$. Specifically, it evaluates \eqref{eq:manipulable} with $\ccalN_t$ (the set of known free cells reachable by the robot at time $t$) replaced by the cells reachable from $\bbp_{\text{rand}}$ on the corresponding simulated map [line~\ref{alg3:mani}, Alg.~\ref{alg:expand}]. The planner then samples which manipulable object to relocate and where to place it. Specifically, an object $o_i\in\ccalO^{M}_{\text{rand}}$ and a candidate configuration $\bbc_i'\in\ccalC_i$ are sampled according to $f_{\ccalO^{M}}$ and $f_{\ccalC_i}$, respectively [lines~\ref{alg3:picko}--\ref{alg3:pickc}, Alg.~\ref{alg:expand}], where
\begin{equation}\label{eq:fo}
    f_{\ccalO^{M}}
    =p_{\text{obs}}\,\delta_{o_{\text{LLM}}}
    +(1-p_{\text{obs}})\,\mathrm{Unif}(\ccalO^{M}_{\text{rand}}),
\end{equation}
\begin{equation}\label{eq:fc}
    f_{\ccalC_i}
    =p_{\text{loc}}\,\mathrm{Unif}\big(\ccalC_i^{r}\big)
    +(1-p_{\text{loc}})\,\mathrm{Unif}\big(\ccalC_i^{\ccalF}\big).
\end{equation}
Here, $\delta_{o_{\text{LLM}}}$ assigns probability one to an object $o_{\text{LLM}}\in\ccalO^{M}_{\text{rand}}$ recommended by a pre-trained LLM; $\ccalC_i^{\ccalF}\subseteq\ccalC_i$ contains the admissible placements of $o_i$ in known free space of the simulated map, with the cells currently occupied by $o_i$ treated as free; and $\ccalC_i^{r}\subseteq\ccalC_i^{\ccalF}$ contains those placements within a user-specified distance $r$ of the current configuration of $o_i$. Thus, \eqref{eq:fo} biases object selection toward the LLM recommendation, whereas \eqref{eq:fc} biases placement sampling toward nearby configurations, thereby favoring shorter relocations. The parameters $p_{\text{obs}},p_{\text{loc}}\in[0,1)$ control the
corresponding biases.
If $\ccalC_i^{r}=\emptyset$, the local component defaults to $\mathrm{Unif}(\ccalC_i^{\ccalF})$. Since $p_{\text{obs}},p_{\text{loc}}<1$, both distributions retain a uniform component over their respective sampling spaces. The LLM prompt used to select $o_{\text{LLM}}$ is detailed in the Appendix. 

The sampled object $o_i$ and configuration $\bbc_i'$ define a candidate carry action that relocates $o_i$ from its configuration $\bbc_i$ in $\bbc_{\text{rand}}$ to $\bbc_i'$, and the manipulation location $j$ remains to be determined [line~\ref{alg3:check}, Alg.~\ref{alg:expand}]. 
The action $a^{j}_{\mathrm{car}}(\bbc_i,\bbc_i')$ is feasible at
$\bbq_{\text{rand}}$ if (i) the grasp cell $m^j(\bbc_i)$ is reachable
from $\bbp_{\text{rand}}$ through known free space; (ii)
$\ccalB_i(\bbc_i')$ lies in known free space; and (iii) a transport path
from $m^j(\bbc_i)$ to $m^j(\bbc_i')$ exists, with the cells vacated by
$o_i$ treated as free, where (ii) and (iii) are the corresponding conditions of Section~\ref{sec:pf}.
\textcolor{black}{Conditions (i) and (iii) are verified by running $A^\star$ over the known free space of the simulated map of $\bbq_{\text{rand}}$, from $\bbp_{\text{rand}}$ to $m^j(\bbc_i)$ and from $m^j(\bbc_i)$ to $m^j(\bbc_i')$, respectively; the two resulting paths are stored with the action so that Alg.~\ref{alg:framework} can execute it without replanning (Section~\ref{sec:decision}).}
%
The manipulation locations are examined in a random order, and the first one satisfying these requirements is selected as $j$ [line~\ref{alg3:newp}, Alg.~\ref{alg:expand}].

If the action is feasible, we construct a new node $\bbq_{\text{new}}=[\bbp',\bbc']$ that is reached after applying that action to $\bbq_{\text{rand}}$: the simulated free space is updated by vacating $\ccalB_i(\bbc_i)$ and blocking $\ccalB_i(\bbc_i')$ [line~\ref{alg3:newfree}, Alg.~\ref{alg:expand}], the object configuration $\bbc'$ is obtained from $\bbc_{\text{rand}}$ by replacing the entry of $o_i$ with $\bbc_i'$ [line~\ref{alg3:newc}, Alg.~\ref{alg:expand}], and the robot state $\bbp'$ is set to the release cell $\bbp'=m^j(\bbc_i')$ of the certified manipulation location [line~\ref{alg3:newp}, Alg.~\ref{alg:expand}]. The new node and the action, together with its stored paths, are returned to Alg.~\ref{alg:planner} [line~\ref{alg3:ret}, Alg.~\ref{alg:expand}], which adds them to the tree [lines~\ref{alg2:if}--\ref{alg2:add}, Alg.~\ref{alg:planner}], and $z(\bbq_{\text{new}})$ is computed for the termination test [line~\ref{alg2:frontier}, Alg.~\ref{alg:planner}]. If infeasible, no node is added [line~\ref{alg3:else}, Alg.~\ref{alg:expand}].

\begin{algorithm}[t]
\footnotesize
\caption{\texttt{TreeExpansion}($\bbq_{\text{rand}}$)}
\LinesNumbered
\label{alg:expand}
Compute the manipulable set $\ccalO^{M}_{\text{rand}}$ on the simulated map of $\bbq_{\text{rand}}$\; \label{alg3:mani}
Sample an object $o_i\in\ccalO^{M}_{\text{rand}}$ from $f_{\ccalO^{M}}$\; \label{alg3:picko}
Sample a configuration $\bbc_i'\in\ccalC_i$ from $f_{\ccalC_i}$\; \label{alg3:pickc}
\eIf{$\exists\, j\in\{1,\dots,J_i\}$ such that $m^j(\bbc_i)$ is reachable from $\bbp_{\text{rand}}$ and conditions (ii)--(iii) of Section~\ref{sec:pf} hold for $\textcolor{black}{a^{j}_{\mathrm{car}}(\bbc_i,\bbc_i')}$ on the simulated map \label{alg3:check}}{
Update the simulated free space: vacate $\ccalB_i(\bbc_i)$, block $\ccalB_i(\bbc_i')$\; \label{alg3:newfree}
$\bbc'\gets\bbc_{\text{rand}}$ with the entry of $o_i$ replaced by $\bbc_i'$\; \label{alg3:newc}
\textcolor{black}{$j \leftarrow$ first manipulation point passing the check;}~ $\bbp'\gets m^j(\bbc_i')$\; \label{alg3:newp}
\Return $\bbq_{\text{new}}=\textcolor{black}{[\bbp',\bbc']}$,~ $\textcolor{black}{a^{j}_{\mathrm{car}}(\bbc_i,\bbc_i')}$ \label{alg3:ret}
}{
\Return $\bbq_{\text{new}}=\varnothing$,~ $a_{\mathrm{car}}=\varnothing$ \label{alg3:else}
}
\end{algorithm}

\textbf{Termination Criterion:}
NAMO-LLM terminates when the hypothesized relocations establish a collision-free path to the goal \cite{11204512}. This criterion is insufficient in partially known environments, where the goal region may lie in unknown space and successful relocations may instead only establish access to the frontier of the clearing path $\pi_{t}^{\text{man}}$ used to evaluate \eqref{eq:decision}. We therefore terminate Alg. \ref{alg:planner} as soon as a node $\bbq$ satisfying
$z(\bbq)\in\{z_{\mathrm{ref}},\bot\}$ is added to the tree.
Specifically, (a) $z(\bbq_{\text{new}})=z_{\mathrm{ref}}$ indicates that the hypothesized relocations enable the detour path to reach the same frontier as the clearing path, thereby resolving the mismatch that invoked the relocation planner; and (b) $z(\bbq_{\text{new}})=\bot$ indicates that the detour path reaches the goal entirely within known space [line~\ref{alg2:target}, Alg.~\ref{alg:planner}]. 

\textbf{Returning a Relocation Plan:}
The planner returns as soon as an expansion yields a target node $\bbq^\star$. Letting $h$ denote its depth in $\ccalT$, the carry actions $(a_1,\dots,a_h)$ along the branch from $\bbq_0$ to $\bbq^\star$ are extracted by backtracking via $A$ and returned to Alg.~\ref{alg:framework} [line~\ref{alg2:extract}, Alg.~\ref{alg:planner}].

\begin{theorem}[Probabilistic completeness]\label{thm}
Assume that $p_{\text{rand}},p_{\text{obs}},p_{\text{loc}} \in [0,1)$, and suppose that the sampling loop of Alg.~\ref{alg:planner} (i.e., lines~\ref{alg2:while}--\ref{alg2:extract}) runs for $n_{\max}$ iterations. If there exists a finite feasible sequence of carry actions $(a_1,\dots,a_h)$ whose final node $\bbq_h$ satisfies $z(\bbq_h)\in\{z_{\mathrm{ref}},\bot\}$, then the probability that Alg.~\ref{alg:planner} returns such a sequence converges to one as $n_{\max}\to\infty$.
\end{theorem}

We note that probabilistic completeness is defined relative to the frozen snapshot. The proof follows the inductive argument of~\cite{van2010path,11204512} and relies on the observation that the distributions~\eqref{eq:fv}--\eqref{eq:fc} retain uniform components and thus sample every feasible action at every node with positive probability. Details are omitted due to space limitations.

\vspace{-0.1cm}
\section{Experiments}\label{sec:sims}
\vspace{-0.1cm}

In this section, we evaluate Alg.~\ref{alg:framework} under various configurations and compare it against a method adapted from \cite{ellis2023navigation}, demonstrating fewer relocations, shorter travel, and lower execution time. All experiments were conducted on a laptop with an AMD Ryzen 9 8945HS 4.0 GHz processor and 16 GB of RAM. Our implementation is available at \cite{codeNAMOLLM_u}.

\vspace{-0.2cm}
\subsection{Setting Up Comparative Experiments}\label{setup}
\vspace{-0.1cm}

\textbf{Environment:} We consider two $16\times12$ m house environments as in Fig.~\ref{fig:intro}. They contain walls, non-movable furniture, and $n$ movable boxes with footprint $0.055\times0.30$ m and two manipulation cells located $0.45$ m beyond the midpoints of their short sides. The robot is modeled as a point after inflating all obstacles by its radius of $0.3$ m and moves at $0.1$ m/s.

\textbf{Sensing and Mapping:} The map of Section~\ref{sec:pf} is constructed at $0.02$ m resolution using a simulated sensor with a $3.5$ m range and $360^\circ$ FoV. Visible cells, computed by recursive shadowcasting, are labeled as free or occupied; all others remain unknown. A box
and its class and configuration are discovered once its center cell is observed.

\textbf{Planner Setup:} We denote our method by $\text{NAMO-LLM}_{\text{u}}(p_{\text{rand}},p_{\text{obs}})$ and evaluate $p_{\text{rand}},p_{\text{obs}}\in\{0.2,0.8\}$ and the extreme variants $(0,0)$ and $(1,1)$. The $(0,0)$ variant uses uniform node and object sampling, removing LLM guidance, and retains the completeness property of Theorem~\ref{thm}; we refer to it as RandomTree. The $(1,1)$ variant always
uses the LLM to select the node to expand and object to relocate 
and does not retain this property; we refer to it as LLM-Plan. We use GPT-5.4 with temperature $1$ throughout. Unless otherwise specified, $p_{\text{loc}}=0.9$, $r=4$ m, and $\delta=5$ s.

\begin{table}[t]
\centering
\footnotesize\setlength{\tabcolsep}{1pt}
\begin{tabular}{|c|c|c|c|c|}
\hline
& Method & $H$ & Dist. (m) & Exec. (Plan.) Time (s) \\ \noalign{\hrule height 1pt}
\multirow{7}{*}{Case \Romannum{1}}
& NAMO-SA$_{\text{u}}$ \cite{ellis2023navigation} & 22.4 & 101.2 & 1053.2(41.1)  \\ \cline{2-5}
& RandomTree & 7.5 & 72.1 & 800.6(80.1)  \\ \cline{2-5}
& NAMO-LLM$_{\text{u}}$(0.2, 0.2) & 6.0 & 57.6 & 638.6(62.6) \\ \cline{2-5}
& NAMO-LLM$_{\text{u}}$(0.8, 0.2) & 7.1 & 64.3 & 707.2(65.1) \\ \cline{2-5}
& NAMO-LLM$_{\text{u}}$(0.2, 0.8) & \textbf{3.9} & \textbf{41.0} & \textbf{458.7}(49.5) \\ \cline{2-5}
& NAMO-LLM$_{\text{u}}$(0.8, 0.8) & 4.4 & 42.9 & 480.3(51.4) \\ \cline{2-5}
& LLM-Plan & 4.6 & 43.2 & 489.1(57.2) \\ \noalign{\hrule height 1pt}
\multirow{7}{*}{Case \Romannum{2}}
& NAMO-SA$_{\text{u}}$ \cite{ellis2023navigation} & 35.2 & 148.8 & 1561.1(73.4) \\ \cline{2-5}
& RandomTree & 10.7 & 88.4 & 1099.5(215.5)  \\ \cline{2-5}
& NAMO-LLM$_{\text{u}}$(0.2, 0.2) & 8.5 & 66.7 & 769.3(102.3) \\ \cline{2-5}
& NAMO-LLM$_{\text{u}}$(0.8, 0.2) & 10.0 & 79.9 & 888.2(89.7) \\ \cline{2-5}
& NAMO-LLM$_{\text{u}}$(0.2, 0.8) & \textbf{5.4} & \textbf{48.6} & \textbf{555.4}(61.1) \\ \cline{2-5}
& NAMO-LLM$_{\text{u}}$(0.8, 0.8) & 5.8 & 49.9 & 562.4(62.4) \\ \cline{2-5}
& LLM-Plan & 6.9 & 53.4 & 609.8(76.0) \\ \noalign{\hrule height 1pt}
\multirow{7}{*}{Case \Romannum{3}}
& NAMO-SA$_{\text{u}}$ \cite{ellis2023navigation} & 24.3 & 154.7 & 1604.2(57.8) \\ \cline{2-5}
& RandomTree & 11.2 & 159.4 & 1774.7(181.0)  \\ \cline{2-5}
& NAMO-LLM$_{\text{u}}$(0.2, 0.2) & 8.8 & 124.2 & 1390.9(150.2) \\ \cline{2-5}
& NAMO-LLM$_{\text{u}}$(0.8, 0.2) & 10.4 & 139.3 & 1530.0(137.7) \\ \cline{2-5}
& NAMO-LLM$_{\text{u}}$(0.2, 0.8) & \textbf{5.8} & \textbf{71.9} & \textbf{830.7}(111.3) \\ \cline{2-5}
& NAMO-LLM$_{\text{u}}$(0.8, 0.8) & 6.4 & 81.3 & 920.7(107.7) \\ \cline{2-5}
& LLM-Plan & 6.7 & 76.1 & 870.9(109.7) \\ \noalign{\hrule height 1pt}
\multirow{7}{*}{Case \Romannum{4}}
& NAMO-SA$_{\text{u}}$ \cite{ellis2023navigation} & 45.3 & 236.2 & 2454.0(90.8) \\ \cline{2-5}
& RandomTree & 16.7 & 221.2 & 2642.4(430.7) \\ \cline{2-5}
& NAMO-LLM$_{\text{u}}$(0.2, 0.2) & 16.3 & 193.6 & 2190.3(254.1) \\ \cline{2-5}
& NAMO-LLM$_{\text{u}}$(0.8, 0.2) & 13.2 & 159.9 & 1887.7(288.8) \\ \cline{2-5}
& NAMO-LLM$_{\text{u}}$(0.2, 0.8) & \textbf{9.9} & \textbf{106.1} & \textbf{1318.1}(258.3) \\ \cline{2-5}
& NAMO-LLM$_{\text{u}}$(0.8, 0.8) & 11.1 & 118.5 & 1419.9(234.3) \\ \cline{2-5}
& LLM-Plan & 11.9 & 115.9 & 1396.3(237.6) \\ \hline
\end{tabular}
\caption{Summary of Results for Case Studies \Romannum{1}--\Romannum{4}.}
\label{tab:comp}
\vspace{-0.5cm}
\end{table}

\textbf{Baseline:} We compare our method against $\text{NAMO-SA}_{\text{u}}$, a search-based method built upon \cite{ellis2023navigation}. The setting in \cite{ellis2023navigation} differs from ours in three ways.
First, it assumes a known static map and discovers only movable objects online; thus, its detour paths can only be blocked by undiscovered movable objects, whereas
ours may also be invalidated by fixed obstacles in unexplored space. Second, as discussed in Section~\ref{sec:intro}, it cannot handle joint relocations in which several manipulable objects must be relocated before any path opens.
Third, it relocates objects to a predefined storage zone, whereas our method samples their placements online.
For a fair comparison, we adapt \cite{ellis2023navigation} to address these differences while preserving its greedy relocation strategy. First, we retain its decision rule but evaluate whether a path to the goal exists on the partially known map by treating unknown cells as free and discovered movable objects as obstacles, i.e., using the detour path $\pi_t^{\text{nav}}$ of Section~\ref{sec:decision}. Since navigation reveals the map, this check is repeated
: if $\pi_t^{\text{nav}}$ exists, the robot follows it for $\delta$ time steps, as in \textsc{Navigate}; otherwise, it relocates the first manipulable object crossed by $\pi_t^{\text{man}}$ (see Section~\ref{sec:decision}), which either opens a path or exposes a manipulation cell of another blocking object; this replaces the cost-based object selection of \cite{ellis2023navigation}. Since the decision is re-evaluated as soon as that object is relocated, the objects along the clearing path are relocated one after another until a path opens. This also allows it to handle joint relocations (see e.g., Cases~\Romannum{1}--\Romannum{4} in Section \ref{sec:comp}). 
Second, instead of a storage zone, the placement of the relocated object is sampled from \eqref{eq:fc} with the same $p_{\text{loc}}$ and $r$ as $\text{NAMO-LLM}_{\text{u}}$, executing the first sample that yields a feasible carry action, i.e., satisfies conditions (i)--(iii) of Section~\ref{sec:pf}. The robot fails if the clearing path does not exist. Thus, the baseline shares our map, placement strategy, navigation interval $\delta$, and feasibility checks, but relocates greedily only when no detour exists rather than searching over relocation sequences.


\textbf{Evaluation Metrics:} \textcolor{black}{We evaluate (a) number of executed relocations $H$; (b) total robot travel distance, including navigation, object approach, and transport; and (c) total execution time, comprising robot travel and planner computation time, where the latter includes map updates, path queries, navigation and relocation planning, and LLM inference. Values in parentheses in Table~\ref{tab:comp} report the planner computation time in seconds. Results are averaged over $100$ runs.}

\vspace{-0.2cm}
\subsection{Comparative Experiments
}\label{sec:comp}
\vspace{-0.1cm}
\textcolor{black}{We consider four case studies in the two environments of Fig.~\ref{fig:intro}, differing in $n$ and relocation dependencies. All require joint relocations, while Cases~II and IV introduce an access dependency. We denote by
$H_{\min}$ the minimum number of relocations needed to reach $\ccalV_g$ on the true map. Table~\ref{tab:comp} summarizes the results.}

\textbf{Case Study \Romannum{1}} ($n=30$, $H_{\min}=3$): 
\textcolor{black}{Our method performs best with $(0.2,0.8)$, relocating $3.9$ boxes on average, close to $H_{\min}$. $\text{NAMO-SA}_{\text{u}}$ 
relocates
$5.7\times$ more boxes and travels $2.5\times$ farther. Its planner computation time is the lowest, as it performs no search, but its execution time is $2.3\times$ longer due to greater travel. Compared to $(0.2,0.8)$, RandomTree
requires $1.9\times$ more relocations and $1.7\times$ longer execution time, and even more planner computation time despite no LLM inference, as more iterations are needed to find a solution; LLM-Plan requires $1.2\times$ more relocations and $7\%$ longer execution time.}

\textbf{Case Study \Romannum{2}} ($n=40$, $H_{\min}=4$):
\textcolor{black}{This case extends Case Study~\Romannum{1} with $10$ additional movable obstacles introducing an access dependency and increasing $H_{\min}$ to $4$.
All $\text{NAMO-LLM}_{\text{u}}$ configurations again outperform $\text{NAMO-SA}_{\text{u}}$. The performance gap with $\text{NAMO-SA}_{\text{u}}$ widens in this larger case with an access dependency, with $(0.2,0.8)$ again performing best. We observe that the LLM makes more mistakes in this more complex environment. Unlike LLM-Plan, $\text{NAMO-LLM}_{\text{u}}$ with $p_{\text{rand}},p_{\text{obs}}<1$ can recover from them through the uniform sampling.}

\textcolor{black}{\textbf{Case Studies \Romannum{3}--\Romannum{4}}
($n=30$, $H_{\min}=4$ and $n=40$, $H_{\min}=5$, respectively):
These cases consider a more complex environment than Cases~\Romannum{1}--\Romannum{2}, with more pathways to the goal. Case~\Romannum{4} adds $10$ movable obstacles introducing an access dependency and increasing $H_{\min}$ to $5$. We observe trends similar to Cases~\Romannum{1}--\Romannum{2}, with $(0.2,0.8)$ performing best, but longer travel and planning times due to greater clutter.}

\begin{rem}[Effect of $p_{\text{loc}}$]
\textcolor{black}{We use Case Study~\Romannum{1} to evaluate the placement-sampling bias in \eqref{eq:fc}. We fix $p_{\text{rand}}=p_{\text{obs}}=0.8$ and vary
$p_{\text{loc}}\in\{0,0.5,0.9\}$, where $p_{\text{loc}}=0$ corresponds to uniform sampling over the known free space. For
$p_{\text{loc}}=0,0.5,0.9$, respectively, we obtain
$H=4.3,4.8,4.4$, travel distances of $55.7,51.5,42.9$ m, and execution times of $616.0$ ($58.5$), $570.7$ ($55.9$), and $480.3$ s ($51.4$), where the values in parentheses are planning times.}
\end{rem}

\vspace{-0.1cm}
\subsection{Validation with a Metric-Semantic Mapping System}\label{sec:hydra}
\vspace{-0.1cm}

\begin{figure}[t]
\centering
\includegraphics[width=0.7\linewidth]{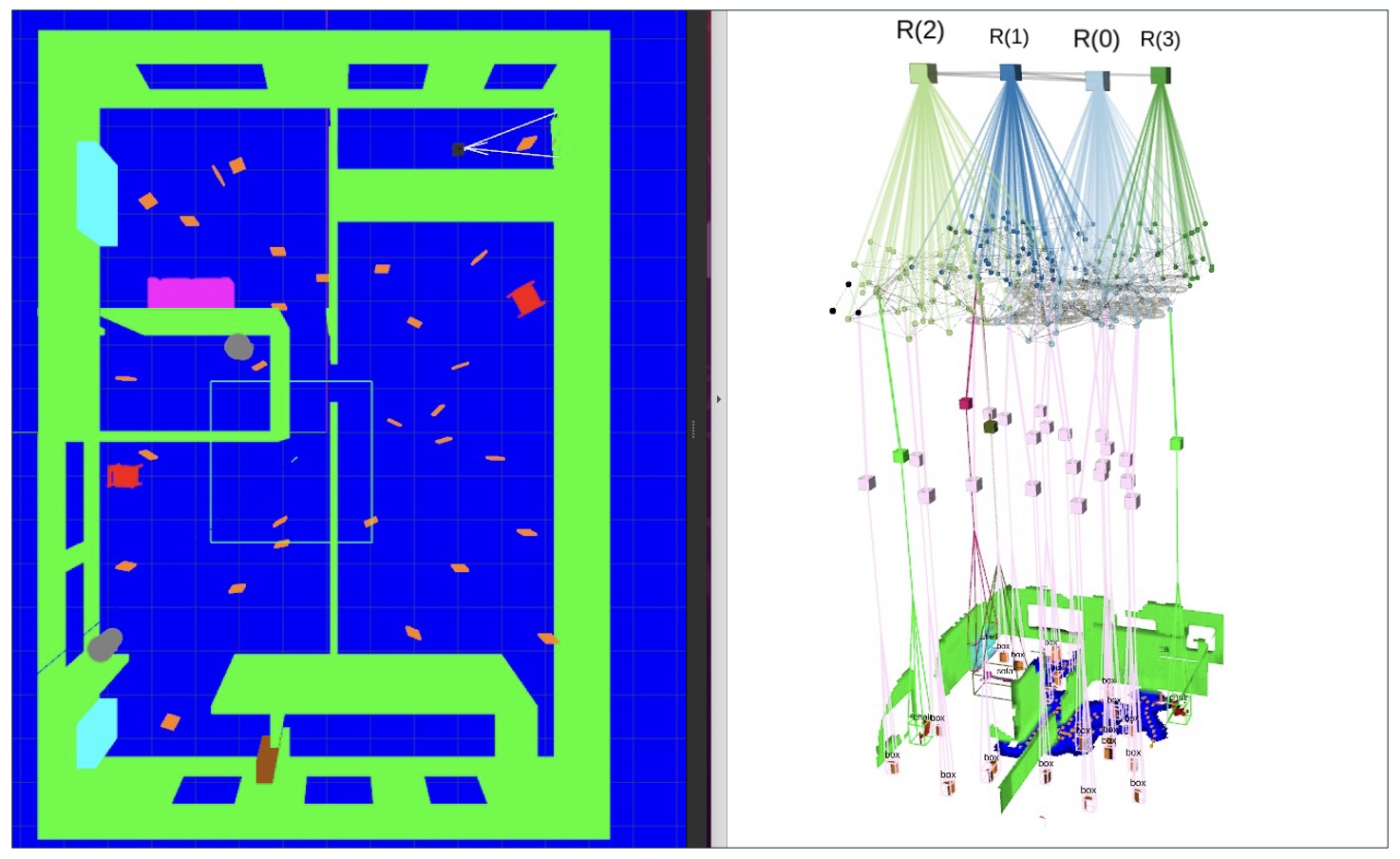}\vspace{-0.3cm}
\caption{Evaluation with Hydra in Gazebo. Left: top-down view of the ground-truth environment of Case Study I. Right: the 3D scene graph built online by Hydra from the robot’s RGB-D stream.}\vspace{-0.2cm}
\label{fig:hydra}
\end{figure}

\textcolor{black}{We also evaluate the framework with realistic perception,
where Assumption~\ref{as:perception} no longer holds. We replace the simulated mapping system of Section~\ref{setup} with Hydra \cite{hughes2022hydra} in ROS~1 Noetic and Gazebo~11; see Fig.~\ref{fig:hydra}.} The environment is identical to Case Study~\Romannum{1}. The robot is a TurtleBot3 Waffle Pi with an OpenMANIPULATOR-X arm, an RGB-D camera ($640\times480$, $80^\circ$ FoV, $10$ m range), and a 2D LiDAR, and uses ground-truth localization. Hydra builds a 3D scene graph from the RGB-D stream using color-based semantic labels, from which we extract the occupancy and object layers of Section~\ref{sec:pf} on a $0.05$ m grid.
\textcolor{black}{
Hydra perceives the unexplored environment, while changes caused by manipulation are incorporated explicitly. After each relocation, the object's footprint is moved from its original to its new pose in both layers, and Hydra's accumulated evidence at these cells is reset. This avoids
waiting for repeated observations to override previously fused measurements.
}
 
We ran $\text{NAMO-LLM}_{\text{u}}(0.8,0.8)$ for $25$ trials, with the robot reaching the goal in $84\%$ of them. Failures result from the perception pipeline: (i) errors in object positions estimated by Hydra prevent the robot from reaching the correct manipulation cell; and (ii) closely spaced objects are occasionally merged into a single object.

\vspace{-0.1cm}
\section{Conclusions}\label{sec:conclusion}

We proposed an online NAMO planner for unknown environments, where the robot decides whether to relocate discovered objects or navigate through unexplored space. Our experiments showed fewer relocations, shorter travel, and lower execution time than baseline methods due to its non-uniform sampling strategy. 
Future work will focus on extensions to 3D environments, hardware validation, and enhancing robustness to perceptual errors.

\vspace{-0.1cm}
\appendix
\section{LLM-guided Object Selection}\label{app:llm}

In this appendix, we discuss how $\ccalV_{\ccalT}^{\text{max}}$ in \eqref{eq:fv} is constructed and  how we prompt a pre-trained LLM to select $o_{\text{LLM}}$ in \eqref{eq:fo}.

\textbf{Node Score:} 
Following \cite{11204512}, each node $\bbq\in\ccalV_{\ccalT}$ is assigned a score $v(\bbq)$ equal to the number of edges along the branch from the root to $\bbq$ that were generated by relocating the LLM-recommended object. The set
$\ccalV_{\ccalT}^{\max}=\{\bbq\in\ccalV_{\ccalT}\mid v(\bbq)=\max_{\bbq'\in\ccalV_{\ccalT}}v(\bbq')\}$ collects the nodes with the highest score and is used to bias the node sampling in \eqref{eq:fv}. The score is updated incrementally as new nodes are added to the tree.

\textbf{LLM Prompt:} The mechanism for incorporating the LLM recommendation into the sampling distributions is inherited from \cite{11204512}. However, the prompt used therein assumes complete knowledge of the environment and cannot be directly applied in our setting: the goal may lie in unexplored space, and neither the geometry nor the objects in that space are available to the planner. We therefore redesign the prompt to provide the LLM only with information available on the simulated map induced by $\bbq_{\mathrm{rand}}$ and to elicit recommendations that remain useful under partial environment knowledge.
Specifically, the simulated map is represented as an ASCII occupancy grid distinguishing known-free, occupied, and unknown cells, together with the robot and goal locations and the discovered movable objects; no ground-truth information about unexplored space is provided. The candidate objects are restricted to $\ccalO^{M}_{\mathrm{rand}}$, ensuring that the recommended object $o_{\mathrm{LLM}}$ can be used directly in \eqref{eq:fo}. Moreover, rather than asking which object should be relocated to clear a path to the goal as in \cite{11204512}, the prompt asks which relocation makes the most progress toward the goal or toward access to unknown space when the goal cannot yet be reached within known space. The prompt used in our experiments is in \cite{codeNAMOLLM_u}.


\textbf{Output:}
The LLM is prompted to return a candidate object $o_{\text{LLM}}$ in a prescribed format. If the output is invalid or identifies an object outside the provided candidate list $\ccalO^{M}_{\text{rand}}$, the recommendation is discarded and $p_{\text{obs}}$ is set to zero for that iteration, reducing \eqref{eq:fo} to uniform sampling over $\ccalO^{M}_{\text{rand}}$.

\vspace{-0.1cm}
\bibliographystyle{IEEEtran}
\bibliography{ref}

\end{document}